\documentclass{article}
\usepackage{iclr2025_conference,times}

\usepackage{amsmath,amssymb}
\usepackage{booktabs}
\usepackage{graphicx}
\usepackage{microtype}
\usepackage{placeins}
\usepackage{tabularx}
\usepackage{xcolor}
\usepackage{xspace}
\usepackage{tikz}
\usetikzlibrary{arrows.meta,backgrounds,calc,fit,positioning}

\definecolor{neoblue}{RGB}{31,119,180}
\definecolor{vizorange}{RGB}{255,127,14}
\definecolor{alignpurple}{RGB}{148,103,189}
\definecolor{policygreen}{RGB}{44,160,44}
\definecolor{frozengray}{RGB}{127,127,127}

\newlength{\figarchnatwidth}
\newcommand{\sref}[1]{Sec.~\ref{#1}}
\newcommand{\pref}[1]{App.~\ref{#1}}

\newcommand{\method}{4DGS-WAM\xspace}

\title{\method: Bridging Past and Future with an Object-Centric World Action Model based on 4D Gaussian Splatting}
\author{
Yueen Ma \\
The Chinese University of Hong Kong \\
Shanghai Academy of AI for Science \\
\texttt{mayueen@link.cuhk.edu.hk}
\And
Zenglin Xu \\
Fudan University \\
Shanghai Academy of AI for Science \\
\texttt{zenglinxu@fudan.edu.cn}
\And
Irwin King \\
The Chinese University of Hong Kong \\
\texttt{king@cse.cuhk.edu.hk}
}
\iclrfinalcopy

\usepackage{hyperref}
\usepackage{url}
\hypersetup{
  hidelinks,
  pdftitle={4DGS-WAM: Bridging Past and Future with an Object-Centric World Action Model based on 4D Gaussian Splatting},
  pdfauthor={Yueen Ma, Zenglin Xu, Irwin King}
}

\begin{document}
\raggedbottom

\maketitle
\lhead{Preprint}
\makeatletter
\begingroup
\long\def\@makefntext#1{\parindent 1em\noindent #1}
\footnotetext{This is a work in progress.}
\endgroup
\makeatother

\begin{abstract}
Current world action models (WAMs) typically operate on 2D visual data. These models can achieve exceptional visual quality, but they lack explicit spatial structure for individual objects and repeatedly process redundant background content. Although point clouds can represent the world in 3D space, they can be difficult to align and accumulate across viewpoints. In this paper, we leverage an explicit 4D Gaussian Splatting (4DGS) representation that separately models dynamic objects and the static background of a scene. For dynamic objects, we use a policy model to predict future actor actions and a world model to predict transformations of their observed Gaussian splats. The static background need not be regenerated for future states, as much of it has already been observed in past frames. This forms an object-centric world action model, which we name 4DGS-WAM. It lifts 2D observations into a persistent 4D representation so that previously observed static content can be reused during future prediction. Future-state extrapolation can then focus on modeling the evolution of dynamic objects. Experiments on KITTI-MOT evaluate short-horizon prediction and past reconstruction.
\end{abstract}

\section{Introduction}

Predicting how the world evolves in response to an agent's actions is a fundamental capability for embodied AI. World action models (WAMs)~\citep{wang2026wam} aim to equip policies with world-modeling capabilities, enabling embodied agents to reason about the consequences of their actions before taking them in the real world. Recent WAMs have achieved impressive visual fidelity by operating directly on sequences of 2D images~\citep{cen2025worldvla,zhang2025epona}. However, modeling the world entirely in image space presents a fundamental challenge: consecutive frames contain substantial redundant information, particularly in static backgrounds, while the underlying 3D structure and motion of dynamic objects remain implicit. As a result, image-based models must repeatedly generate large portions of the scene that are unchanged or only weakly related to the agent's actions.

An explicit spatial representation offers a promising alternative. By reconstructing a scene in 4D, a model can associate observations from different viewpoints and represent dynamic scenes persistently rather than regenerating them frame by frame. Point clouds~\citep{huang2026pointworld} provide one such representation, but their sparse and unstructured nature can make dense view synthesis and temporal accumulation difficult, especially under viewpoint changes and object motion.

In this work, we propose \method, a world action model built on an explicit 4D Gaussian Splatting (4DGS)~\citep{wu2024_4dgs} representation. Our central idea is to decompose a dynamic scene into dynamic and static components and model them separately. The dynamic component represents objects whose states evolve over time, while the static component captures persistent background content observed over time. Rather than predicting an entire future image, we use a policy model to predict the future actions of dynamic objects and a world model to predict the corresponding transformations of their Gaussian splats. Because much of the background visible in future observations has already been observed in previous frames, it need not be reconstructed from scratch at every predicted time step. We then compose the extrapolated dynamic component with the accumulated static component to represent the future scene.

To achieve dynamic--static decomposition, we leverage a suite of vision foundation models (VFMs)~\citep{awais2025vfm} for segmentation, depth estimation, and camera pose estimation. These estimates allow us to lift observations into a common world coordinate system and track the motion of individual objects over time. We further use an optical-flow VFM to provide pixel-level motion cues, enabling the modeling of non-rigid object transformations. In contrast to prior WAMs built on multimodal large language model (MLLM) backbones~\citep{cen2025worldvla,wang2026univla}, \method adopts a fundamentally different framework: a policy network and a world model operating on top of the perceived 4DGS representation. The policy network takes object-centric trajectories and a queried horizon $h$ and predicts target-horizon actor actions. Conditioned on these actions and the same $h$, the world model predicts transformations of the Gaussian splats associated with each object. This object-centric formulation grounds world action modeling in the explicit spatial structure of the scene.

We apply \method to autonomous driving tasks. On the KITTI-MOT~\citep{geiger2012kitti} benchmark, rendering the accumulated 4DGS state from observed camera views achieves higher reported full-frame metrics than 3DGS~\citep{kerbl2023gs}, point-cloud methods, and other 4DGS baselines. For future extrapolation, we compare \method against video- and 4DGS-based world models. A future application of \method is robot manipulation, where the model must handle more challenging physical interactions, such as object collisions and deformable objects.

\begin{figure*}[t]
\centering
\input{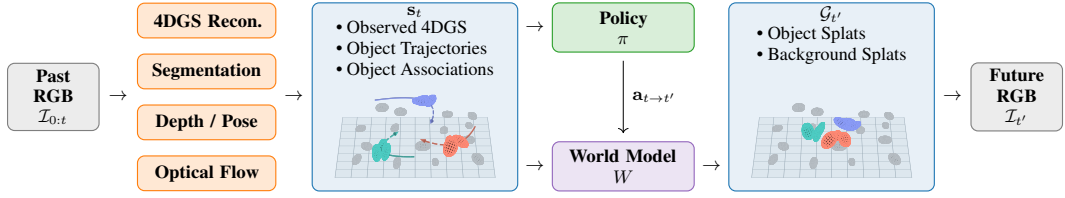}
\caption{\textbf{Architecture of \method{}.} Past RGB frames $\mathcal{I}_{0:t}$ are processed by a 4DGS reconstruction module and vision foundation models for segmentation, depth, camera pose, and optical flow. Policy~$\pi$ predicts target-horizon $\mathrm{SE}(3)$ actions $\widehat{\mathbf{a}}_{t\rightarrow t'}$, which condition $W$ to produce the future 4DGS state for rendering.}
\label{fig:architecture}
\end{figure*}

In summary, our contributions are:
\begin{itemize}
    \item We introduce \method, a 4D Gaussian Splatting-based world action model that explicitly decomposes dynamic scenes into static and dynamic components.
    
    \item We propose a WAM framework in which policy and world models operate on object trajectories and Gaussians derived from a scene-level 4DGS state, distinct from prior multimodal LLM-based WAMs.
    
    \item We evaluate \method on an autonomous driving benchmark, KITTI-MOT, comparing future prediction with video- and 4DGS-based world models and reconstruction with 3D and 4D mapping methods.
\end{itemize}
\section{Related Work}
\label{sec:related}

\paragraph{World action models.}
Recent world action models~\citep{wang2026wam} primarily use image- or video-based representations. WorldVLA~\citep{cen2025worldvla} and RynnVLA-002~\citep{cen2025rynnvla002} jointly model future images and actions in unified autoregressive frameworks, while UniVLA~\citep{wang2026univla} incorporates world modeling into a token-based vision-language-action model. DreamZero~\citep{ye2026dreamzero} jointly generates future video and actions using video diffusion. In driving, GAIA-1~\citep{hu2023gaia1}, DriveDreamer-2~\citep{zhao2025drivedreamer2}, and Epona~\citep{zhang2025epona} generate future driving video, with Epona additionally modeling future trajectories. PointWorld~\citep{huang2026pointworld} instead predicts action-conditioned 3D point flows. In contrast, \method{} maintains a renderable 4D Gaussian state with an explicit static--dynamic decomposition and applies a time-conditioned policy to object trajectories and a world model to object Gaussians.

\begin{figure*}[t]
\centering
\setlength{\abovecaptionskip}{2pt}%
\begingroup
\def\g{0.32cm}
\def\gs{0.72cm}
\def\yFut{4.53cm}

\ifdefined\wampanelA\else
  \newsavebox{\wampanelA}\newsavebox{\wampanelB}%
\fi
\ifdefined\wamscaledA\else
  \newsavebox{\wamscaledA}\newsavebox{\wamscaledB}%
\fi

\sbox{\wampanelB}{%
\begin{tikzpicture}[
  font=\footnotesize,
  block/.style 2 args={
    draw=#1, fill=#2, rounded corners=3.5pt, align=center,
    minimum height=0.65cm, text width=2.20cm, inner sep=3pt, line width=0.55pt
  },
  reconblock/.style 2 args={
    draw=#1, fill=#2, rounded corners=3.5pt, align=center,
    minimum height=0.68cm, text width=3.05cm, inner sep=3pt, line width=0.55pt
  },
  port/.style={draw=frozengray!50, circle, minimum size=0.78cm,
    inner sep=0pt, font=\footnotesize\itshape, text=frozengray,
    line width=0.55pt},
  arrow/.style={->, line width=0.4pt, shorten <=1pt, shorten >=1pt,
    rounded corners=3pt}
]
  \node[port] (rgbin) at (0,0.35) {$\mathcal{I}_{0:t}$};
  \node[below=2pt of rgbin] (labrgbin) {RGB};
  \node[reconblock={vizorange}{vizorange!18}, above=\g of rgbin] (recon)
    {\textbf{4DGS Perception}};
  \node[port, above=\g of recon] (state) {$\mathbf{s}_t$};
  \node[port] (bact) at ([yshift={\gs+0.325cm}]state.center) {$\mathbf{a}_{t\rightarrow t'}$};
  \node[block={policygreen}{policygreen!18}, left=\g of bact] (policy)
    {\textbf{Policy}\\$\pi$};
  \node[block={alignpurple}{alignpurple!18}, right=\g of bact] (world)
    {\textbf{World Model}\\$W$};
  \node[port, above=\g of world] (futurestate) {$\mathcal{G}_{t'}$};
  \node[above=1pt of futurestate] (labfut) {4DGS};
  \node[port, right=\g of futurestate] (rgbout) {$\mathcal{I}_{t'}$};
  \node[above=1pt of rgbout] (labrgbout) {RGB};
  \draw[arrow] (rgbin) -- (recon);
  \draw[arrow] (recon) -- (state);
  \draw[arrow] (state.west) -| (policy.south);
  \draw[arrow] (state.east) -| (world.south);
  \draw[arrow] (policy) -- (bact);
  \draw[arrow] (bact) -- (world);
  \draw[arrow] (world) -- (futurestate);
  \draw[arrow] (futurestate) -- (rgbout);
\end{tikzpicture}%
}

\sbox{\wampanelA}{%
\begin{tikzpicture}[
  font=\footnotesize,
  wamblock/.style 2 args={
    draw=#1, fill=#2, rounded corners=3.5pt, align=center,
    inner sep=2pt, line width=0.55pt
  },
  iodblock/.style 2 args={
    draw=#1, fill=#2, rounded corners=3.5pt, align=center,
    minimum height=0.65cm, minimum width=3.05cm, inner sep=3pt,
    line width=0.55pt, font=\footnotesize
  },
  port/.style={draw=frozengray!50, circle, minimum size=0.78cm,
    inner sep=0pt, font=\footnotesize\itshape, text=frozengray,
    line width=0.55pt},
  arrow/.style={->, line width=0.4pt, shorten <=1pt, shorten >=1pt,
    rounded corners=3pt}
]
  \node[port] (obs) at (1.525,0.35) {$\mathcal{I}_t$};
  \node[port] (actionin) at (4.895,0.35) {$\mathbf{a}_t$};
  \node[below=2pt of obs] (labobs) {RGB};
  \node[below=2pt of actionin] {Action};
  \node[iodblock={frozengray}{frozengray!12}, above=\g of obs] (imgenc)
    {\textbf{Image Encoder}};
  \node[iodblock={frozengray}{frozengray!12}, above=\g of actionin] (actenc)
    {\textbf{Action Encoder}};
  \node[port] (actionout) at (1.525,\yFut) {$\mathbf{a}_{t'}$};
  \node[port] (nextobs) at (4.895,\yFut) {$\mathcal{I}_{t'}$};
  \node[above=1pt of actionout] (labactout) {Action};
  \node[above=1pt of nextobs] {RGB};
  \node[iodblock={frozengray}{frozengray!12}, below=\g of actionout] (actdec)
    {\textbf{Action Decoder}};
  \node[iodblock={frozengray}{frozengray!12}, below=\g of nextobs] (imgdec)
    {\textbf{Image Decoder}};
  \coordinate (encmid) at ($(imgenc.north)!0.5!(actenc.north)$);
  \path let
    \p1=(imgenc.north),
    \p2=(actdec.south),
    \p3=(imgenc.west),
    \p4=(actenc.east),
    \n1={\y2-\y1-2*\g},
    \n2={\x4-\x3}
  in node[wamblock={frozengray}{frozengray!15}, minimum height=\n1,
    minimum width=\n2, above=\g of encmid] (vwam) {\textbf{WAM Backbone}};
  \draw[arrow] (obs) -- (imgenc);
  \draw[arrow] (actionin) -- (actenc);
  \draw[arrow] (imgenc.north) -- (vwam.south -| imgenc.north);
  \draw[arrow] (actenc.north) -- (vwam.south -| actenc.north);
  \draw[arrow] (vwam.north -| actdec.south) -- (actdec.south);
  \draw[arrow] (vwam.north -| imgdec.south) -- (imgdec.south);
  \draw[arrow] (actdec) -- (actionout);
  \draw[arrow] (imgdec) -- (nextobs);
\end{tikzpicture}%
}

\pgfmathsetmacro{\wamscale}{\linewidth/\figarchnatwidth}
\sbox{\wamscaledA}{\scalebox{\wamscale}{\usebox{\wampanelA}}}%
\sbox{\wamscaledB}{\scalebox{\wamscale}{\usebox{\wampanelB}}}%
\ifdefined\wamHt\else\newdimen\wamHt\fi
\wamHt=\ht\wamscaledB
\advance\wamHt\dp\wamscaledB
\sbox{\wamscaledA}{\resizebox{\wd\wamscaledA}{\wamHt}{\usebox{\wamscaledA}}}%
\sbox{\wamscaledB}{\resizebox{\wd\wamscaledB}{\wamHt}{\usebox{\wamscaledB}}}%

\begin{tikzpicture}
  \pgfmathsetlengthmacro{\wamgap}{0.5*(0.5*\linewidth-max(\wd\wamscaledA,\wd\wamscaledB))}
  \coordinate (div) at (0.5\linewidth, 0);
  \node[inner sep=0, outer sep=0, anchor=east] (na)
    at ([xshift=-\wamgap]div) {\usebox{\wamscaledA}};
  \node[inner sep=0, outer sep=0, anchor=west] (nb)
    at ([xshift=\wamgap]div) {\usebox{\wamscaledB}};
  \node[anchor=north, inner sep=1pt] (capa) at ([yshift=-4pt]na.south) {(a)};
  \node[anchor=north, inner sep=1pt] (capb) at ([yshift=-4pt]nb.south) {(b)};
  \draw[densely dashed, draw=frozengray!70, line width=0.4pt]
    (div |- na.south) -- (div |- na.north);
  \useasboundingbox (0, 0 |- capa.south) rectangle (\linewidth, 0 |- na.north);
\end{tikzpicture}
\endgroup
\caption{\textbf{World action model comparison.} \textbf{(a) Representative autoregressive WAMs.} Image and action tokens are predicted jointly in a unified model. \textbf{(b) \method{}.} Our model maintains a decomposed, renderable 4D Gaussian state: the static bank persists while policy-conditioned world dynamics transport only dynamic object splats.}
\label{fig:wam-comparison}
\end{figure*}
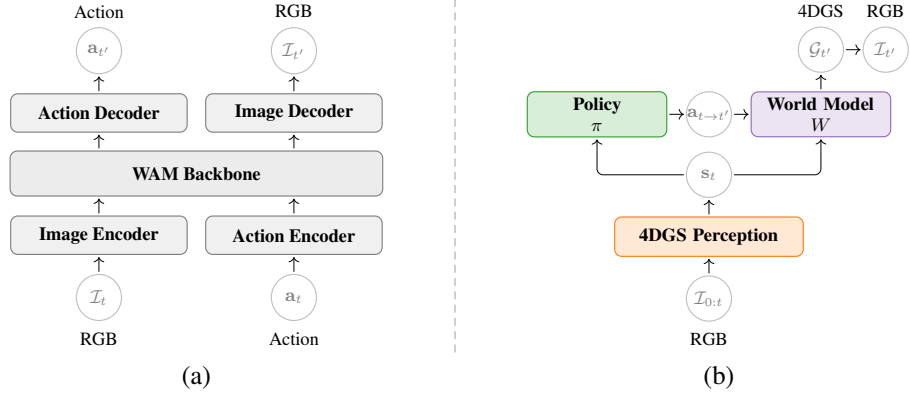

\paragraph{4D Gaussian Splatting.} 3D Gaussian Splatting (3DGS)~\citep{kerbl2023gs} provides an explicit and efficient radiance-field representation for static scenes. 4DGS-SLAM~\citep{li2025_4dgsslam} and Flow4DGS-SLAM~\citep{wang2026flow4dgsslam} extend Gaussian splatting to dynamic SLAM, but do not model action-conditioned future states. NeoVerse~\citep{yang2026neoverse} performs pose-free, feed-forward 4D reconstruction from monocular video and supports novel-trajectory video generation. DriveDreamer4D~\citep{zhao2025drivedreamer4d} uses a video world model to synthesize novel-trajectory driving videos and then optimizes 4DGS using aligned real and synthetic views, rather than directly predicting future 4DGS states. Envision4D~\citep{song2026envision4d} performs feed-forward 4DGS future extrapolation for driving, but is neither object-centric nor action-conditioned and does not decompose persistent static and evolving dynamic content.

\paragraph{Vision foundation models.} Vision foundation models~\citep{awais2025vfm} provide complementary object-centric cues for our method. SAM~3~\citep{carion2026sam3} supplies segmentation masks and object identities over time, while DA3~\citep{lin2026da3} estimates dense depth maps; VGGT~\citep{wang2025vggt} and VGGT-$\Omega$~\citep{wang2026vggtomega} additionally recover camera geometry. WAFT~\citep{wang2026waft} estimates consecutive-frame optical flow. D4RT~\citep{zhang2026d4rt} is a unified alternative for dynamic 4D reconstruction and tracking.

\section{Method}
\label{sec:method}

\method first reconstructs past observations into an object-centric 4DGS state, then uses a time-conditioned policy and a time- and action-conditioned world model to extrapolate that state and render future observations, as illustrated in Figure~\ref{fig:architecture}.

\begin{figure*}[t]
\centering
\setlength{\abovecaptionskip}{6pt}%
\begingroup
\resizebox{0.79\linewidth}{!}{%
\begin{tikzpicture}[
  font=\footnotesize,
  modwide/.style 2 args={
    draw=#1, fill=#2, rounded corners=3.5pt, align=center,
    minimum height=0.65cm, minimum width=3.20cm, inner sep=3pt,
    line width=0.55pt
  },
  modattn/.style 2 args={
    draw=#1, fill=#2, rounded corners=3.5pt, align=center,
    minimum height=0.65cm, minimum width=3.45cm, text width=3.25cm,
    inner sep=3pt, line width=0.55pt
  },
  modhead/.style 2 args={
    draw=#1, fill=#2, rounded corners=3.5pt, align=center,
    minimum height=0.65cm, minimum width=2.80cm, inner sep=3pt,
    line width=0.55pt
  },
  port/.style={draw=frozengray!50, circle, minimum size=0.78cm,
    inner sep=0pt, font=\footnotesize\itshape, text=frozengray,
    line width=0.55pt},
  arrow/.style={->, line width=0.4pt, shorten <=1pt, shorten >=1pt,
    rounded corners=3pt}
]
  \def\short{0.22cm}
  \def\px{1.15}
  \def\actsym{\mathbf{a}_{t\rightarrow t'}}

  \begin{scope}[yshift=1.20cm]
  \def\coresep{5pt}
  \node[port] (hist) at (\px-1.90,0.0) {$\mathcal{T}_{0:t}$};
  \node[port] (memin) at (\px,0.0) {$\mathbf{m}$};
  \node[port] (hpi) at (\px+1.90,0.0) {$h$};
  \node[below=2pt of hist] (labhist) {Trajectories};
  \node[below=2pt of memin] {Memory};
  \node[below=2pt of hpi] {Horizon};

  \coordinate (intop) at (\px,0 |- hist.north);
  \node[modwide={frozengray}{frozengray!15}, anchor=south]
    (embed) at ([yshift=\short]intop) {\textbf{Embedding}};

  \node[modattn={policygreen}{policygreen!18}, anchor=south]
    (temp) at ([yshift={\short+\coresep}]embed.north)
    {\textbf{Temporal Attn.\ $\times L$}};
  \node[modattn={policygreen}{policygreen!18}, above=3pt of temp]
    (interact) {\textbf{Actor $k$NN Attention}};
  \begin{scope}[on background layer]
    \node[draw=policygreen, fill=policygreen!5, rounded corners=3.5pt,
      line width=0.8pt, inner sep=\coresep, fit=(temp)(interact)]
      (picore) {};
  \end{scope}

  \node[modwide={policygreen}{policygreen!18}, anchor=south]
    (head) at ([yshift=\short]picore.north) {\textbf{Action Head}};
  \node[port, anchor=south]
    (action) at ([yshift=\short]head.north) {$\actsym$};
  \node[above=2pt of action] {Action};

  \draw[arrow] (hist.north) -- (embed.south west);
  \draw[arrow] (memin.north) -- (embed.south -| memin);
  \draw[arrow] (hpi.north) -- (embed.south east);
  \draw[arrow] (embed.north) -- (picore.south);
  \draw[arrow] (picore.north) -- (head.south);
  \draw[arrow] (head.north) -- (action.south);
  \end{scope}

  \def\bx{10.4}
  \def\col{1.70}
  \def\lx{\bx-\col}
  \def\rx{\bx+\col}
  \def\wcoresep{5pt}
  \def\turnrise{0.22cm}

  \def\hin{0.85}
  \pgfmathsetmacro{\xain}{\lx-1.85}
  \pgfmathsetmacro{\xsplats}{\bx-\hin}
  \pgfmathsetmacro{\xhw}{\bx+\hin}
  \node[port] (ain) at (\xain,0.0) {$\actsym$};
  \node[port] (splats) at (\xsplats,0.0) {$\mathcal{G}_t$};
  \node[port] (hw) at (\xhw,0.0) {$h$};
  \node[below=2pt of ain] (labsplats) {Action};
  \node[below=2pt of splats] {Splats};
  \node[below=2pt of hw] {Horizon};

  \coordinate (wintop) at (\bx,0 |- splats.north);
  \node[modwide={frozengray}{frozengray!15}, anchor=south]
    (sembed) at ([yshift=\short]wintop) {\textbf{Splat Embedding}};
  \node[modwide={frozengray}{frozengray!15}, above=3pt of sembed]
    (knn) {\textbf{Part Tokens}};

  \node[modwide={alignpurple}{alignpurple!18}, anchor=south]
    (eattn) at ([yshift={\short+\wcoresep}]knn.north)
    {\textbf{Edge Attention}};
  \node[modwide={alignpurple}{alignpurple!18}, above=3pt of eattn]
    (nmlp) {\textbf{SwiGLU + FiLM}};
  \coordinate (wmidE) at ($(eattn.east)!0.5!(nmlp.east)$);
  \coordinate (wmidW) at ($(eattn.west)!0.5!(nmlp.west)$);
  \node[font=\small\bfseries, text=alignpurple, inner sep=0pt,
    right=3pt of wmidE] (xl) {$\times L$};
  \node[font=\small\bfseries, inner sep=0pt,
    left=3pt of wmidW] (xlph) {\phantom{$\times L$}};
  \begin{scope}[on background layer]
    \node[draw=alignpurple, fill=alignpurple!5, rounded corners=3.5pt,
      line width=0.8pt, inner sep=\wcoresep, fit=(eattn)(nmlp)(xl)(xlph)]
      (wcore) {};
  \end{scope}

  \node[modhead={alignpurple}{alignpurple!18}, anchor=south]
    (geo) at ([yshift=\short]wcore.north -| {\lx,0}) {\textbf{Geometry Head}};
  \node[modhead={alignpurple}{alignpurple!18}, anchor=south]
    (app) at ([yshift=\short]wcore.north -| {\rx,0}) {\textbf{Appearance Head}};
  \node[modhead={frozengray}{frozengray!15}, above=3pt of geo]
    (compose) {\textbf{Compose}};
  \node[modhead={frozengray}{frozengray!15}, above=3pt of app]
    (reweight) {\textbf{Residual Mix}};

  \coordinate (fbase) at ({\bx},0 |- compose.north);
  \node[port, anchor=south]
    (future) at ([yshift=\turnrise]fbase)
    {$\mathcal{G}_{t'}$};
  \node[above=2pt of future] {Future Splats};

  \draw[arrow] (splats.north) -- (sembed.south -| splats);
  \draw[arrow] (hw.north) -- (sembed.south -| hw);
  \draw[arrow] (ain.north) |- (wcore.west);
  \draw[arrow] (ain.north) |- (compose.west);
  \draw[arrow] (knn.north) -- (wcore.south -| knn);
  \draw[arrow] (wcore.north -| geo) -- (geo.south);
  \draw[arrow] (wcore.north -| app) -- (app.south);
  \draw[arrow] (compose.north) |- (future.west);
  \draw[arrow] (reweight.north) |- (future.east);

  \path let \p1=(labhist.south), \p2=(labsplats.south),
    \n1={min(\y1,\y2)} in coordinate (capbase) at (0,\n1);
  \node[anchor=north] at ([yshift=-2pt]\px,0 |- capbase)
    {(a) Policy~$\pi$};
  \node[anchor=north] at ([yshift=-2pt]\bx,0 |- capbase)
    {(b) World Model~$W$};

  \coordinate (div) at ($(picore.east)!0.5!(ain.west)$);
  \draw[densely dashed, draw=frozengray!70, line width=0.4pt]
    (div |- capbase) -- (div |- future.north);
\end{tikzpicture}%
}
\endgroup
\caption{\textbf{Architectures of the policy network and the world model.} \textbf{(a)~Policy network~$\pi$.} Each actor trajectory is augmented with a learned memory token $\mathbf{m}$ and a horizon embedding of the queried $h$, and the resulting sequences are processed by stacked temporal-attention layers followed by actor-level $k$NN attention. The action head reads $\mathbf{m}$ and predicts a target-horizon $\mathrm{SE}(3)$ action. \textbf{(b)~World model~$W$.} Conditioned on the same $h$ and the predicted action, splat embeddings and category-specific part tokens are processed by stacked geometric tensor layers. Residual $\mathrm{SE}(3)$ twists predicted at the part nodes are blended onto splats, and the resulting transformations are composed with the object action; source appearance attributes are updated through residual mixing.}
\label{fig:policy-world-modules}
\end{figure*}
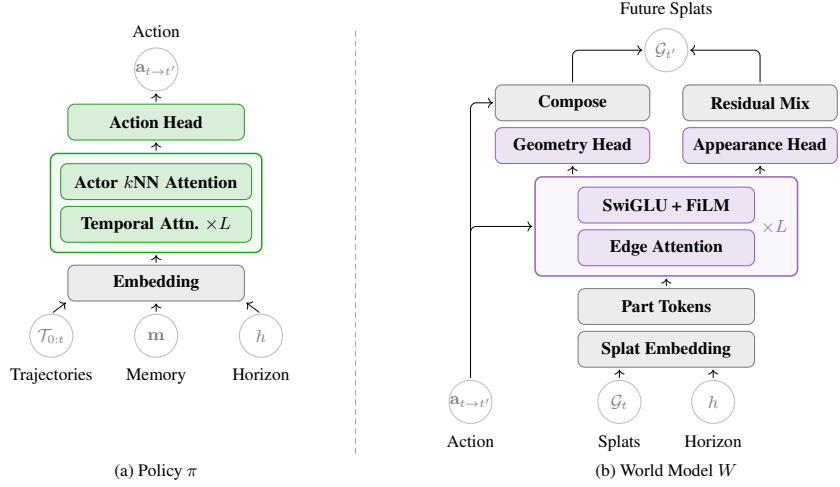

\subsection{Past Perception}
\label{sec:past}

Given past RGB frames $\mathcal{I}_{0:t}$, a 4DGS reconstructor~\citep{yang2026neoverse} produces $N$ Gaussian primitives:
\begin{equation}
  \mathcal{G}_{0:t}=
  \left\{\left(
  \boldsymbol{\mu}_i,\boldsymbol{r}_i,\boldsymbol{sc}_i,
  \alpha_i,\boldsymbol{sh}_i,\boldsymbol{\tau}_i,
  \boldsymbol{\xi}^{+}_i,\boldsymbol{\xi}^{-}_i
  \right)\right\}_{i=1}^{N},
  \label{eq:4dgs}
\end{equation}
where each primitive contains the standard 3DGS attributes---mean position $\boldsymbol{\mu}_i$, rotation $\boldsymbol{r}_i$, scale $\boldsymbol{sc}_i$, opacity $\alpha_i$, and spherical-harmonic appearance coefficients $\boldsymbol{sh}_i$---together with a temporal lifespan $\boldsymbol{\tau}_i=(\tau_i^{\mathrm{start}},\tau_i^{\mathrm{end}})$ and forward and backward motion twists $\boldsymbol{\xi}^{+}_i$ and $\boldsymbol{\xi}^{-}_i$. These twists characterize each Gaussian's reconstructed trajectory over its lifespan within the observed frames. We denote by $\mathcal{G}_t$ the active Gaussian snapshot obtained by evaluating $\mathcal{G}_{0:t}$ at time $t$.

\paragraph{Vision foundation models.} 4DGS reconstruction alone lacks object-centric information and therefore cannot distinguish dynamic objects from the static background. We use vision foundation models (VFMs) to estimate segmentation, optical flow, depth, and camera poses. Segmentation provides an object mask $\mathcal{M}^{(o)}_n$, defined as the set of pixels $p$ belonging to object $o$ at time $n$. These masks support object tracking, while optical flow supplies pixel-level motion cues. We use depth maps and camera poses to lift these image-level estimates into 3D object trajectories and world-space 3D flow correspondences in a common coordinate frame. For each dynamic object $o\in\mathcal{O}_t$, we denote its object-center trajectory up to time $t$ and its 3D flow correspondences from time $t$ to a target time $t'$ by
\begin{equation}
  \mathcal{T}^{(o)}_{0:t}=
  \left\{\mathbf{c}^{(o)}_n\in\mathbb{R}^3\right\}_{n=0}^{t},
  \qquad
  \mathcal{F}^{(o)}_{t\rightarrow t'}=
  \left\{\bigl(\mathbf{x}_j,\mathbf{f}_j\bigr)\right\}_{j\in\mathcal{J}^{(o)}_{t\rightarrow t'}},
  \label{eq:object-motion-observations}
\end{equation}
where $\mathbf{c}^{(o)}_n$ is the center of object $o$, $\mathbf{x}_j$ is the world-space source position of a matched splat-associated surface point at time $t$, and $\mathbf{f}_j$ is its measured world-space displacement to time $t'$. Object centers and flow correspondences are lifted using segmentation masks, optical flow, depth, and camera poses. During training, target-time masks, centers, and correspondences are extracted from available future observations; at inference, no target-time quantities are used. The masks at time $t$ associate active Gaussians with dynamic objects, yielding $\{\mathcal{G}^{(o)}_t\}_{o\in\mathcal{O}_t}$, while the remaining Gaussians form $\mathcal{G}_{\mathrm{static}}$. We define the abstract state $\mathbf{s}_t$ as the reconstructed 4DGS, its object associations, and the object-center trajectories. The object trajectories are provided to the policy, while the flow correspondences supervise the world model.

\subsection{Future Prediction}
\label{sec:future}

\paragraph{Policy.} A time-conditioned policy network $\pi(\mathbf{s}_t,h)$ predicts actor motions for a queried future time $t'=t+h$, where $h>0$ is the prediction horizon. Let $\mathcal{A}_t=\mathcal{O}_t\cup\{\mathrm{ego}\}$ denote the policy's actor set. The policy receives the center trajectories of these actors and outputs one six-vector target-horizon action $\widehat{\mathbf{a}}^{(o)}_{t\rightarrow t'}\in\mathbb{R}^{6}$ per actor. For any queried $h$, including $h>1$, this action is obtained in a single policy evaluation of the observed prefix, without autoregressive prediction of intermediate actions. The policy produces
\begin{equation}
  \left\{\widehat{\mathbf{a}}^{(o)}_{t\rightarrow t'}\right\}_{o\in\mathcal{A}_t}
  =
  \pi\!\left(
  \left\{\mathcal{T}^{(o)}_{0:t}\right\}_{o\in\mathcal{A}_t},
  h\right).
  \label{eq:time-policy}
\end{equation}
Because the ego camera is mounted on the ego vehicle, $\pi$ includes the ego trajectory in its actor set; its predicted action determines the future camera pose. As shown in Figure~\ref{fig:policy-world-modules}, we append a learned memory token $\mathbf{m}$ to each actor trajectory and add a horizon embedding to every token. The resulting sequences are processed by $L$ causal temporal-attention blocks~\citep{vaswani2017attention}, allowing each memory token to attend to its full trajectory while preventing trajectory tokens from attending to it. Actor-level $k$NN attention then enables interactions among nearby actors. The action head reads only from the corresponding memory token.

We train the policy with Charbonnier regression~\citep{charbonnier1994two}:
\begin{equation}
  \mathcal{L}_{\pi}=
  \frac{1}{|\mathcal{A}_t|}
  \sum_{o\in\mathcal{A}_t}\rho_{\pi}\!\left(
  \widehat{\mathbf{a}}^{(o)}_{t\rightarrow t'}
  -
  \mathbf{a}^{(o)}_{t\rightarrow t'}\right),
  \qquad
  \rho_{\pi}(\mathbf{z})=
  \sqrt{\sum_d z_d^2+\epsilon^2}.
  \label{eq:policy-loss}
\end{equation}
Here, $\mathbf{a}^{(o)}_{t\rightarrow t'}$ is the teacher target-horizon action. Details of the action representation and exact objective are provided in~\pref{app:arch}.

\paragraph{World model.} A time- and action-conditioned world model $W$ predicts the future Gaussian state at prediction horizon $h$. Given target-horizon actions, $W$ operates on the observed, object-associated Gaussian snapshot $\mathcal{G}_t$; we omit the association metadata in the notation below. Because the representation is 4DGS, the perceived Gaussians support novel-view synthesis. For the static background $\mathcal{G}_{\mathrm{static}}$, we aggregate past multi-view observations and render it at the target camera pose. For each dynamic object, conditioned on $\mathbf{a}^{(o)}_{t\rightarrow t'}$, $W$ predicts residual $\mathrm{SE}(3)$ twists on its category-specific part tokens, blends these twists onto the object's splats, and composes the resulting transformations with the supplied action. These operations update Gaussian means and orientations and apply a residual appearance update. Writing $\mathbf{a}_{t\rightarrow t'}=\{\mathbf{a}^{(o)}_{t\rightarrow t'}\}_{o\in\mathcal{O}_t}$, the state transition is
\begin{equation}
  \begin{aligned}
    \widehat{\mathcal{G}}_{t'}
    &= W\!\left(\mathbf{s}_t,\mathbf{a}_{t\rightarrow t'},h\right)
    = W\!\left(\mathcal{G}_t,\mathbf{a}_{t\rightarrow t'},h\right)
    = W\!\left(
    \left(\bigcup_{o\in\mathcal{O}_t}\mathcal{G}^{(o)}_t\right)
    \cup\mathcal{G}_{\mathrm{static}},
    \mathbf{a}_{t\rightarrow t'},h\right) \\
    &=
    \left(\bigcup_{o\in\mathcal{O}_t}
    W\!\left(\mathcal{G}^{(o)}_t,\mathbf{a}^{(o)}_{t\rightarrow t'},h\right)\right)
    \cup\mathcal{G}_{\mathrm{static}}
    =
    \left(\bigcup_{o\in\mathcal{O}_t}\widehat{\mathcal{G}}^{(o)}_{t'}\right)
    \cup\mathcal{G}_{\mathrm{static}}.
  \end{aligned}
  \label{eq:world-transition}
\end{equation}
Rendering $\widehat{\mathcal{G}}_{t'}$ from the target camera pose yields the future observation $\widehat{\mathcal{I}}_{t'}$. During training, we condition $W$ on teacher actions $\mathbf{a}_{t\rightarrow t'}$ and horizon $h$; at inference, we use policy predictions $\widehat{\mathbf{a}}_{t\rightarrow t'}$. This per-object factorization omits inter-object interactions in $W$ and therefore cannot model collisions. As shown in Figure~\ref{fig:policy-world-modules}, $W$ is a GotenNet-style geometric tensor network~\citep{aykent2025gotennet} operating over splats and part tokens. Architectural details, including category-specific part-token construction, are provided in~\pref{app:arch}.

We supervise geometry with 3D flow transport and appearance with masked photometry:
\begin{equation}
  \begin{aligned}
    \mathcal{L}_{W}
    &=
    \mathcal{L}_{\mathrm{transport}}
    + \lambda_{\mathrm{photo}}\mathcal{L}_{\mathrm{photo}},
    \\
    \mathcal{L}_{\mathrm{transport}}
    &=
    \mathcal{L}_{\mathrm{splat}}
    + \mathcal{L}_{\mathrm{center}}
    \\
    &=
    \mathbb{E}_{\substack{o\in\mathcal{O}_t\\j\in\mathcal{J}^{(o)}_{t\rightarrow t'}}}
    \rho_{W}\!\left(
    \widehat{\mathbf{x}}_j
    - \mathbf{x}_j
    - \mathbf{f}_j\right)
    +
    \mathbb{E}_{o\in\mathcal{O}_t}
    \rho_{W}\!\left(
    \operatorname*{mean}_{j\in\mathcal{J}^{(o)}_{t\rightarrow t'}}
    \bigl(\widehat{\mathbf{x}}_j - \mathbf{x}_j\bigr)
    -
    \mathbf{t}\!\left(\mathbf{a}^{(o)}_{t\rightarrow t'}\right)\right),
    \\
    \mathcal{L}_{\mathrm{photo}}
    &=
    \mathbb{E}_{o\in\mathcal{O}_t,\,p\in\mathcal{M}^{(o)}_{t'}}
    \left\|
    \widehat{\mathcal{I}}_{t'}(p)
    - \mathcal{I}_{t'}(p)
    \right\|_1,
    \qquad
    \rho_{W}(\mathbf{z})
    =
    \sum_d\sqrt{z_d^2+\epsilon^2}.
  \end{aligned}
  \label{eq:world-loss}
\end{equation}
Here, $\widehat{\mathbf{x}}_j$ is the predicted destination of the source point $\mathbf{x}_j$, and $\mathbf{t}(\mathbf{a})$ extracts an action's translation. The splat term matches predicted point displacements to measured 3D flow, while the center term matches their average predicted displacement to the supplied target-horizon teacher translation. The photometric term is an $\ell_1$ error within the target-time segmentation mask $\mathcal{M}^{(o)}_{t'}$. Unlike $\rho_{\pi}$, which applies a joint Charbonnier penalty to the full action residual, $\rho_{W}$ applies Charbonnier penalties independently to each coordinate. Optimizer settings and loss weights are given in~\sref{sec:impl}; remaining training details are provided in~\pref{app:objectives}.

\setcounter{topnumber}{1}
\setcounter{dbltopnumber}{1}
\begin{table*}[t]
\centering
\setlength{\abovecaptionskip}{4pt}
\setlength{\belowcaptionskip}{2pt}
\caption{\textbf{Future prediction} on KITTI-MOT.
Mean${\pm}$std over the three sequence-level means at horizons $h{=}1,3$
($6$ frames per method).
Every method is scored on the same prefixes and target frames.
The best given-camera result in each column is bold; predicted-camera
rows are not eligible for this mark.
$^{\dagger}$DriveDreamer-2 uses KITTI-adapted conditioning in place of its native nuScenes inputs.}
\label{tab:future-prediction}
\setlength{\tabcolsep}{4pt}
\small
\begin{tabular*}{\textwidth}{@{}l@{\extracolsep{\fill}}lccc@{}}
\toprule
Method & Camera & PSNR$\uparrow$ & SSIM$\uparrow$ & LPIPS$\downarrow$ \\
\midrule
Epona & given
  & $17.46{\pm}1.21$ & $0.482{\pm}0.054$ & $0.203{\pm}0.021$ \\
DriveDreamer-2 & given$^{\dagger}$
  & $13.73{\pm}2.67$ & $0.360{\pm}0.082$ & $0.426{\pm}0.064$ \\
Envision4D & predicted
  & $17.19{\pm}0.65$ & $0.514{\pm}0.072$ & $0.307{\pm}0.069$ \\
\method{} (policy) & given
  & $\mathbf{18.80{\pm}1.21}$ & $\mathbf{0.596{\pm}0.091}$ & $\mathbf{0.161{\pm}0.042}$ \\
\method{} (policy) & predicted
  & $16.47{\pm}3.05$ & $0.482{\pm}0.033$ & $0.245{\pm}0.107$ \\
\bottomrule
\end{tabular*}

\end{table*}

\section{Experiments}
\label{sec:experiments}

We evaluate \method{} on the KITTI-MOT~\citep{geiger2012kitti} benchmark
for two tasks: future-observation prediction and past reconstruction of
observed frames.
Figures~\ref{fig:future-prediction-examples}--%
\ref{fig:past-reconstruction-examples} show qualitative examples.

\subsection{Implementation details}
\label{sec:impl}

We train $\pi$ and $W$ separately. The policy is a width-$96$ transformer with $L{=}3$ temporal-attention blocks; the world model is a width-$128$ geometric tensor network with $L{=}4$ layers. The policy uses AdamW~\citep{loshchilov2019adamw} with learning rate $3{\times}10^{-4}$, batch size $16$, and $40$ epochs, with $1.5$ epochs of linear warmup, cosine decay to $0.1$ times the peak rate, and gradient clipping at $5$. The world model uses Adam~\citep{kingma2015adam} with learning rate $3{\times}10^{-5}$ on both geometry and appearance, batch size $4$, and $10$ epochs, with cosine decay and the same gradient clip. Instantiating~\eqref{eq:world-loss}, we weight $\mathcal{L}_{\mathrm{photo}}$, $\mathcal{L}_{\mathrm{splat}}$, and $\mathcal{L}_{\mathrm{center}}$ by $1$, $0.5$, and $0.5$, add a missing-alpha penalty with total weight $0.2$, and add a geometry-parameter $\ell_2$ anchor of weight $0.05$.

\subsection{Future prediction}
\label{sec:future-results}

We compare our future RGB predictions (\sref{sec:future}) with
Epona~\citep{zhang2025epona},
DriveDreamer-2~\citep{zhao2025drivedreamer2}, and
Envision4D~\citep{song2026envision4d}. All methods receive the same observed
prefixes. Epona and DriveDreamer-2 synthesize a future video frame and are
scored at the given future camera; DriveDreamer-2 uses KITTI-adapted
conditioning in place of its native nuScenes inputs. Envision4D jointly
predicts appearance and camera and is scored at that predicted camera.
For \method{}, $\pi$ predicts target-horizon actions and $W$ applies them
to the observed dynamic Gaussians; both models are conditioned on $h$.
We report the same $\pi{\to}W$ prediction at the given future pose and at the
ego camera predicted by $\pi$. The \method{} renders composite predicted
splats over a frozen NeoVerse fusion underlay that fills remaining
uncovered pixels; the baselines are scored on their native outputs.
Table~\ref{tab:future-prediction} lists PSNR,
SSIM~\citep{wang2004ssim}, and LPIPS~\citep{zhang2018lpips}.

At the given camera, policy-fed \method{} leads the video baselines on
all three metrics. At the predicted ego camera, the same prediction trails
Envision4D, which models both content and camera.

\subsection{Past reconstruction}
\label{sec:past-results}

\begin{table*}[t]
\centering
\setlength{\abovecaptionskip}{4pt}
\setlength{\belowcaptionskip}{2pt}
\caption{\textbf{Past reconstruction} of observed frames on KITTI-MOT.
Mean${\pm}$std over three sequences.
The \method{} full-frame scores use a frozen NeoVerse fusion underlay;
other methods are scored on their native outputs. Its dynamic-region
scores come from the corresponding no-fusion evaluation.}
\label{tab:past-reconstruction}
\setlength{\tabcolsep}{3pt}
\small
\begin{tabular*}{\textwidth}{@{}l@{\extracolsep{\fill}}ccccc@{}}
\toprule
Method & PSNR$\uparrow$ & SSIM$\uparrow$ & LPIPS$\downarrow$ & dyn-PSNR$\uparrow$ & dyn-LPIPS$\downarrow$ \\
\midrule
\multicolumn{6}{@{}l}{\textit{Point Cloud Reconstruction}} \\
VGGT-$\Omega$
  & $6.66{\pm}0.17$ & $0.046{\pm}0.017$ & $0.861{\pm}0.038$
  & --- & --- \\
VGGT-SLAM
  & $6.64{\pm}0.39$ & $0.105{\pm}0.051$ & $0.746{\pm}0.070$
  & --- & --- \\
\midrule
\multicolumn{6}{@{}l}{\textit{3DGS Reconstruction}} \\
3DGS
  & $14.54{\pm}2.14$ & $0.477{\pm}0.087$ & $0.525{\pm}0.156$
  & $12.83{\pm}3.72$ & $0.504{\pm}0.142$ \\
MonoGS
  & $20.42{\pm}5.55$ & $0.652{\pm}0.225$ & $0.236{\pm}0.177$
  & $15.51{\pm}5.12$ & $0.224{\pm}0.159$ \\
EmbodiedSplat
  & $13.81{\pm}1.23$ & $0.447{\pm}0.120$ & $0.407{\pm}0.099$
  & $12.23{\pm}0.05$ & $0.411{\pm}0.043$ \\
\midrule
\multicolumn{6}{@{}l}{\textit{4DGS Reconstruction}} \\
4DGS-SLAM
  & $12.09{\pm}1.11$ & $0.446{\pm}0.130$ & $0.462{\pm}0.110$
  & $12.38{\pm}2.66$ & $0.361{\pm}0.093$ \\
Flow4DGS-SLAM
  & $12.69{\pm}0.86$ & $0.502{\pm}0.113$ & $0.400{\pm}0.144$
  & $12.91{\pm}3.10$ & $0.282{\pm}0.100$ \\
\method{}
  & $\mathbf{27.63{\pm}2.07}$ & $\mathbf{0.888{\pm}0.030}$ & $\mathbf{0.053{\pm}0.012}$
  & $\mathbf{25.53{\pm}3.58}$ & $\mathbf{0.025{\pm}0.006}$ \\
\bottomrule
\end{tabular*}

\end{table*}

We compare past reconstructions of the observed prefix (\sref{sec:past}),
rendered at the given past cameras and scored only on those input views,
not mixed with the future-prediction evaluation. Point-cloud baselines
are VGGT-$\Omega$~\citep{wang2026vggtomega} and
VGGT-SLAM~\citep{maggio2025vggtslam}; 3DGS baselines are
3DGS~\citep{kerbl2023gs}, MonoGS~\citep{matsuki2024monogs}, and
EmbodiedSplat~\citep{chhablani2025embodiedsplat}; 4DGS baselines are
4DGS-SLAM~\citep{li2025_4dgsslam} and
Flow4DGS-SLAM~\citep{wang2026flow4dgsslam}. \method{} merges per-frame
static leftovers and dynamic object splats and fills remaining uncovered
pixels with a frozen NeoVerse fusion underlay.
Table~\ref{tab:past-reconstruction} lists PSNR, SSIM, LPIPS, and scores in
dynamic-object regions.

\method{} has the highest reported full-frame and dynamic-region metrics in
Table~\ref{tab:past-reconstruction}. Among 3DGS methods, MonoGS is the
strongest baseline.

\subsection{Qualitative results}
\label{sec:qual}

\begin{figure*}[t]
\centering
\setlength{\abovecaptionskip}{4pt}
\includegraphics[width=\linewidth]{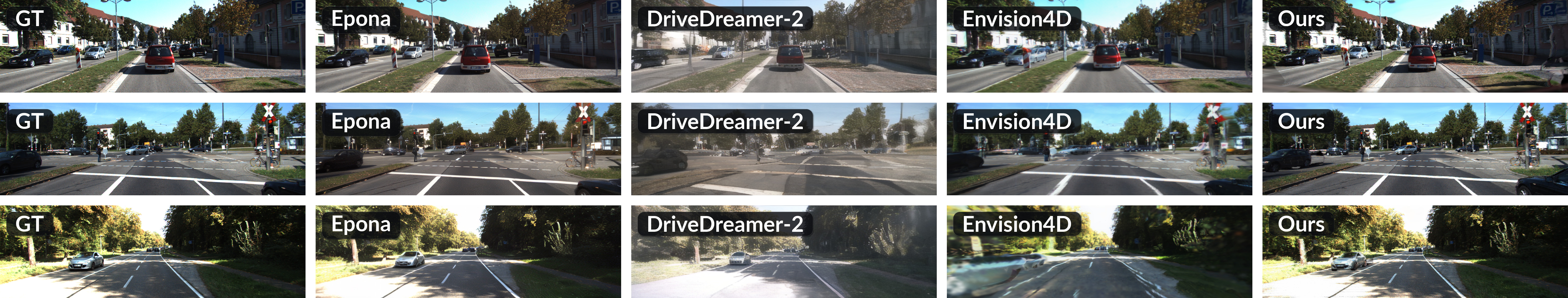}
\caption{\textbf{Future prediction} on KITTI-MOT scenes.
The three rows use horizons $h{=}1$, $h{=}2$, and $h{=}1$, respectively.
Left to right: ground truth, Epona, DriveDreamer-2,
Envision4D, and policy-fed \method{}. Epona and DriveDreamer-2 use the
given future camera; Envision4D and \method{} use their predicted cameras.
The \method{} panels include the frozen fusion underlay described in the text.}
\label{fig:future-prediction-examples}
\vspace{0.6em}
\includegraphics[width=\linewidth]{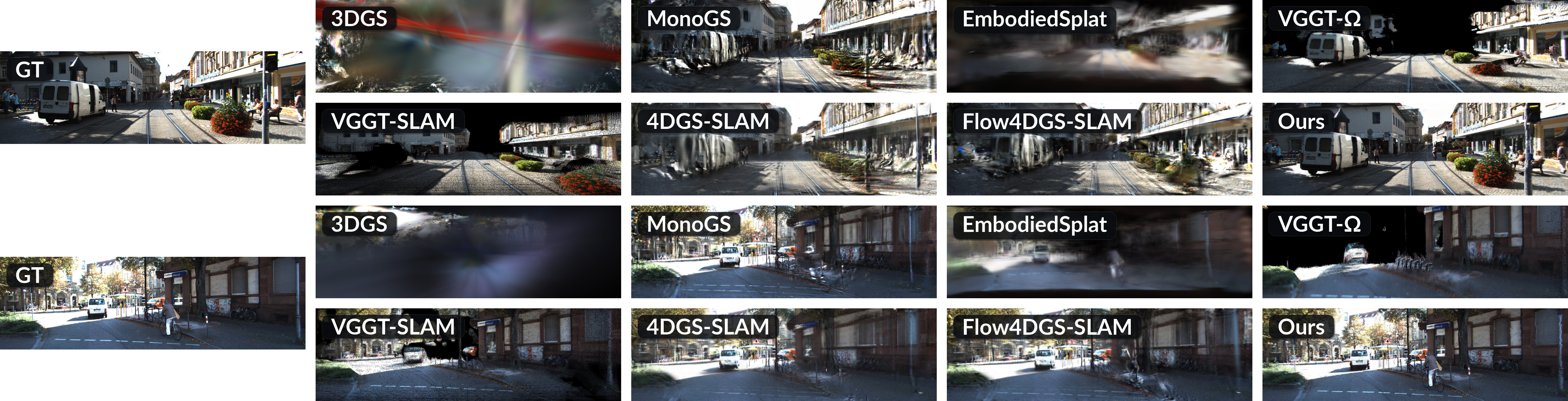}
\caption{\textbf{Past reconstruction} on KITTI-MOT scenes.
Left: ground truth. Right: 3DGS, MonoGS, EmbodiedSplat, and
VGGT-$\Omega$ (upper row); VGGT-SLAM, 4DGS-SLAM, Flow4DGS-SLAM, and
\method{} (lower row). The \method{} panel shows its merged
static--dynamic reconstruction with the frozen fusion underlay.}
\label{fig:past-reconstruction-examples}
\end{figure*}

We visualize KITTI-MOT scenes for future prediction
(Figure~\ref{fig:future-prediction-examples}) and past reconstruction
(Figure~\ref{fig:past-reconstruction-examples}).
Panels share the native image aspect ratio; outputs produced on a different
canvas are resized to that geometry only for display.

For future prediction, each method produces a future RGB frame from the
same observed prefix. Epona and DriveDreamer-2 synthesize that frame as
video and are shown at the given future camera. Envision4D jointly
predicts future appearance and the ego camera. For \method{}, the predicted
ego action sets the render camera so that the panel matches Envision4D's
predicted-camera setting. Its frozen fusion underlay fills remaining
uncovered pixels; the static background is reconstructed from the observed
prefix rather than newly generated.

For past reconstruction, every method maps only the observed prefix and
is rendered at the given past camera, never mixed with future-prediction
views. 3DGS, MonoGS, and EmbodiedSplat optimize a Gaussian scene on
those input views. VGGT-$\Omega$ and VGGT-SLAM lift a point cloud that
we project into the same camera; uncovered pixels stay empty.
4DGS-SLAM and Flow4DGS-SLAM run dynamic Gaussian mapping on the prefix.
\method{} shows the merged static-and-dynamic 4DGS that serves as the
world-model state, together with its frozen fusion underlay, rendered at
the same given camera.

\FloatBarrier
\setcounter{topnumber}{2}
\setcounter{dbltopnumber}{2}

\section{Conclusion}
\label{sec:conclusion}

We presented \method{}, an object-centric world action model on an explicit 4D Gaussian Splatting state. Past observations are reconstructed into static background and dynamic object Gaussians; a time-conditioned policy predicts $\mathrm{SE}(3)$ actor actions, including ego motion, and a time- and action-conditioned world model transports only the dynamic splats from the observed Gaussians, leaving the accumulated background in place. On KITTI-MOT, our merged reconstruction obtains the highest reported full-frame and dynamic-region metrics. Policy-fed short-horizon prediction leads the evaluated video world models at a given camera. By lifting 2D observations into a persistent 4D representation, \method{} reuses reconstructed static content and extrapolates the future through the evolution of dynamic objects.

\section{Limitations and Future Work}
\label{sec:limitations}

The world model factorizes per object and therefore assumes that objects do not collide and cannot instantiate objects that appear after time $t$. The current 4DGS state also has no shape-completion module for unobserved sides of an object, such as the back of a vehicle seen only from the ego camera. These approximations suit the non-contact driving examples studied here, but they are insufficient for robot manipulation, where contact and other inter-object interactions must be modeled. Handling deformable objects, which would support more diverse manipulation tasks in embodied AI, may require more sophisticated 4DGS representations. \method{} also depends on a suite of vision foundation models, so perception errors can propagate into the reconstructed state and predicted dynamics. The static bank preserves observed content but cannot synthesize newly revealed background regions. Our quantitative \method{} renders therefore use a frozen fusion underlay for remaining uncovered pixels, whereas baselines are scored on their native outputs. Prediction is evaluated only at short horizons; longer-horizon rollout remains open. A unified end-to-end model is left to future work. Finally, 4DGS is only one possible 4D representation: the same object-centric WAM could be instantiated on voxel grids, meshes, or other 4D scene encodings.

\FloatBarrier
\bibliographystyle{iclr2025_conference}
\setlength{\bibsep}{2pt}
\bibliography{references}

\clearpage
\appendix

\section{Architecture and Training Details}
\label{app:arch}

\subsection{Notation}
\label{app:notation}

Table~\ref{tab:notation} collects the notation used in the main text and
the implementation details below. Bold lower-case symbols denote vectors,
and calligraphic symbols denote sets or structured states.

\begin{table*}[h!]
\caption{\textbf{Core notation.} The symbols follow the conventions of the
main text.}
\label{tab:notation}
\centering
\footnotesize
\setlength{\tabcolsep}{4pt}
\renewcommand{\arraystretch}{1.12}
\begin{tabularx}{\textwidth}{@{}lX@{}}
\toprule
Symbol & Meaning \\
\midrule
$t,t',h=t'-t$ & Current time, target time, and prediction horizon \\
$\mathcal{I}_n,\widehat{\mathcal{I}}_{t'}$ & RGB observation at time $n$ and rendered future prediction \\
$\mathcal{G}_{0:t},\mathcal{G}_n,\widehat{\mathcal{G}}_{t'}$ & Reconstructed 4DGS, active snapshot, and predicted future state \\
$\mathcal{G}^{(o)}_n,\mathcal{G}_{\mathrm{static}}$ & Object-specific Gaussian subset and static Gaussian bank \\
$\mathbf{s}_t$ & Reconstructed 4DGS, object associations, and center trajectories \\
$\mathcal{O}_t,\mathcal{A}_t$ & Dynamic scene objects and policy actors, where $\mathcal{A}_t=\mathcal{O}_t\cup\{\mathrm{ego}\}$ \\
$\mathcal{M}^{(o)}_n$ & Image mask of object $o$ at time $n$ \\
$\mathcal{T}^{(o)}_{0:t},\mathbf{c}^{(o)}_n$ & Center trajectory and world-space object center \\
$\mathcal{F}^{(o)}_{t\rightarrow t'},\mathcal{J}^{(o)}_{t\rightarrow t'}$ & 3D flow correspondences and their index set \\
$\boldsymbol{\mu}_i$ & Gaussian mean position \\
$\mathbf{x}_j,\widehat{\mathbf{x}}_j,\mathbf{f}_j$ & Geometry-loss source, predicted destination, and measured displacement \\
$\boldsymbol{r}_i,\boldsymbol{sc}_i,\alpha_i$ & Gaussian rotation, scale, and opacity \\
$\boldsymbol{sh}_i,\mathbf{rgb}_i$ & Spherical-harmonic appearance and displayed RGB color \\
$\boldsymbol{\tau}_i,\boldsymbol{\xi}^{+}_i,\boldsymbol{\xi}^{-}_i$ & Gaussian lifespan and reconstructed forward/backward motion twists \\
$\mathbf{a}^{(o)}_{t\rightarrow t'},\widehat{\mathbf{a}}^{(o)}_{t\rightarrow t'}$ & Teacher and predicted target-horizon actions \\
$\pi,W$ & Policy network and Gaussian world model \\
$\mathbf{z}^{(o)}_n,\mathbf{m}$ & 9-D center-state token and learned policy memory token \\
$\mathbf{a}^{(o)}_{\mathrm{CV}},\widehat{\mathbf{u}}^{(o)}_{t,h},\boldsymbol{\sigma}$ & Per-step constant-velocity prior, normalized residual prediction, and residual scales \\
$q_o,\chi_{od}$ & Object confidence and supervised action-dimension mask \\
$g_a=(R_a,\mathbf{t}_a),\boldsymbol{\omega}_a,\theta,\boldsymbol{\iota}_a$ & $\mathrm{SE}(3)$ action transform, rotation vector, angle, and invariant feature triplet \\
$\mathbf{y}_i,\mathbf{p}_k,K$ & Centered splat position, part seed, and number of part tokens \\
$\boldsymbol{\phi}_i,\mathbf{V}_i$ & Invariant scalar features and $C{\times}3$ equivariant vector features \\
$\mathbf{q}_i,\boldsymbol{\Delta}_{ij},d_{ij},\mathbf{e}^{\ell}_{ij},\mathcal{N}(i)$ & Graph-node position, edge geometry, invariant feature, and typed neighborhood \\
$w_{ij}^{m},w_{ik}$ & Edge-attention weight for head $m$ and splat-to-part assignment \\
$\boldsymbol{\delta}_k,\boldsymbol{\delta}_i,g_i$ & Part residual twist, blended splat residual twist, and composed transform \\
$T_{ij},\widetilde{\mathbf{rgb}}_i$ & Local appearance-transport weight and transported source color \\
$\mathbf{d}_i^{\mathrm{rgb}},d_i^{\alpha},\mathbf{d}_i^{\log\mathrm{sc}}$ & Per-splat color, opacity-logit, and log-scale residuals \\
$\mathcal{L}_{\pi},\mathcal{L}_{W}$ & Policy and world-model objectives \\
$\mathcal{L}_{\mathrm{splat}},\mathcal{L}_{\mathrm{center}},\mathcal{L}_{\mathrm{photo}}$ & Dense-flow, center-transport, and masked-photometric losses \\
$\mathcal{L}_{\mathrm{miss}},\mathcal{L}_{\mathrm{anchor}}$ & Missing-alpha penalty and geometry-parameter anchor \\
$\rho_{\pi},\rho_W,\epsilon$ & Joint and coordinatewise Charbonnier penalties and stabilizer \\
\bottomrule
\end{tabularx}
\end{table*}

\subsection{Policy Network}
\label{app:policy}

\paragraph{Action convention.}
The six-vector action order is
\begin{equation}
\mathbf{a}
=
(\Delta x,\Delta y,\Delta z,\mathrm{roll},\mathrm{pitch},\mathrm{yaw}).
\label{eq:action-order}
\end{equation}
It maps to translation
$\mathbf{t}=(\Delta x,\Delta y,\Delta z)$ and rotation
\begin{equation}
R(\mathbf{a})
=R_z(\mathrm{roll})R_x(\mathrm{pitch})R_y(\mathrm{yaw}).
\label{eq:action-rotation}
\end{equation}
Thus, when roll and pitch are zero, the convention reduces to planar yaw
about the vertical axis.

The action head parameterizes the target-horizon prediction as a bounded
residual over a per-step constant-velocity prior:
\begin{equation}
\widehat{\mathbf{a}}^{(o)}_{t\rightarrow t'}
=
\Phi_h\!\left(
\mathbf{a}^{(o)}_{\mathrm{CV}}
+
\boldsymbol{\sigma}\odot
\tanh\!\left(\widehat{\mathbf{u}}^{(o)}_{t,h}\right)
\right),
\label{eq:action-residual}
\end{equation}
where
\begin{equation}
\boldsymbol{\sigma}
=
(0.90,0.10,0.90,0.10,0.10,0.25).
\label{eq:action-scales}
\end{equation}
The first three entries are measured in meters and the last three in
radians. The horizon map $\Phi_h$ converts this per-step-equivalent
parameterization into the cumulative transform from $t$ to $t'$ passed to
the world model: it scales translation by $h$ and sets
$R(h)=\operatorname{Exp}(h\operatorname{Log}R)$. Its inverse divides
translation by $h$ and uses
$\operatorname{Exp}(\operatorname{Log}R/h)$. For object supervision,
target-horizon translation is obtained
from SAM+DA3 centroid displacement and target-horizon yaw from the
tangent-derived heading change; roll and pitch are set to zero and
supervised. For ego supervision, the relative camera motion defines the
target-horizon action.

\paragraph{Input tokens and attention.}
For each actor, the policy uses the latest eight center-state observations,
\begin{equation}
\mathbf{z}^{(o)}_n
=
\left[
\mathbf{c}^{(o)}_n;\,
\mathbf{c}^{(o)}_n-\mathbf{c}^{(o)}_{n-1};\,
\boldsymbol{\nu}^{(o)}_n
\right]
\in\mathbb{R}^{9},
\label{eq:center-token}
\end{equation}
where $\boldsymbol{\nu}^{(o)}_n\in\mathbb{R}^{3}$ is an optional
uncertainty feature and is zero when unavailable. The ego row uses the same
token layout, with centers obtained from the camera trajectory.

The 9-D tokens are projected to width $96$. A learned projection of the
scalar horizon $h$ is added to every history token and to a learned memory
token $\mathbf{m}$. Three causal temporal-attention blocks, each with four
heads, RMS normalization~\citep{zhang2019rmsnorm}, and a SwiGLU
feed-forward layer~\citep{shazeer2020glu}, process the resulting sequence.
Because $\mathbf{m}$ is appended after the history, it
can attend to all valid observations while history tokens cannot attend to
it. A 3D $k$NN interaction layer with $k=4$ then exchanges information
among nearby actors. The normalized memory feature is the sole input to
the action head.

\subsection{World Model}
\label{app:world}

\paragraph{Part-token construction.}
For an object with $N$ active splats, let
\begin{equation}
\mathbf{c}=\frac{1}{N}\sum_{i=1}^{N}\boldsymbol{\mu}_i,
\qquad
\mathbf{y}_i=\boldsymbol{\mu}_i-\mathbf{c}.
\label{eq:centered-splats}
\end{equation}
Farthest-point sampling over $\{\mathbf{y}_i\}$ yields $K$ part seeds
$\{\mathbf{p}_k\}_{k=1}^{K}$. We use $K=1$ for cars, vans, trucks, buses,
trains, and unrecognized categories; $K=3$ for bicycles and motorcycles;
and $K=5$ for people. The $K=1$ case provides a rigid fallback.

\paragraph{Scalar and vector features.}
Let $g_a=(R_a,\mathbf{t}_a)$ be the transform obtained from the supplied
target-horizon action,
$\boldsymbol{\omega}_a=\operatorname{Log}(R_a)$, and
$\theta=\|\boldsymbol{\omega}_a\|_2$. Define the invariant action feature
$\boldsymbol{\iota}_a=[\theta,\|\mathbf{t}_a\|_2,h]$. Splat and part nodes
are initialized as
\begin{align}
\boldsymbol{\phi}^{0}_i
&=
\operatorname{Emb}\!\left(
[\log\boldsymbol{sc}_i;\mathbf{rgb}_i;\alpha_i;h]
\right),
&
\mathbf{V}^{0}_i
&=
\begin{bmatrix}
\mathbf{y}_i^\top\\
\mathbf{t}_a^\top\\
\boldsymbol{\omega}_a^\top
\end{bmatrix},
\label{eq:splat-lift}
\\
\boldsymbol{\phi}^{0}_k
&=
\operatorname{Emb}\!\left(
[\mathbf{0}_3;\mathbf{0}_3;1;h]
\right),
&
\mathbf{V}^{0}_k
&=
\begin{bmatrix}
\mathbf{p}_k^\top\\
\mathbf{t}_a^\top\\
\boldsymbol{\omega}_a^\top
\end{bmatrix}.
\label{eq:part-lift}
\end{align}
The three rows of $\mathbf{V}$ are equivariant vector channels. Action
direction enters these channels directly, while
$\boldsymbol{\iota}_a$ conditions each scalar update through
FiLM~\citep{perez2018film}.

\paragraph{Typed geometric tensor layers.}
The packed graph contains splat--splat $k$NN edges with $k=16$, complete
splat--part edges in both directions, and a complete part--part graph.
For an edge $i\leftarrow j$, define
\begin{equation}
\boldsymbol{\Delta}_{ij}=\mathbf{q}_j-\mathbf{q}_i,
\qquad
d_{ij}=\|\boldsymbol{\Delta}_{ij}\|_2,
\label{eq:edge-geometry}
\end{equation}
where $\mathbf{q}$ denotes either a centered splat or a part seed. The
invariant edge feature is
\begin{equation}
\mathbf{e}^{\ell}_{ij}
=
\left[
d_{ij};\,
\langle\mathbf{V}^{\ell}_i,\mathbf{V}^{\ell}_j\rangle_n;\,
\langle\mathbf{V}^{\ell}_i,\boldsymbol{\Delta}_{ij}\rangle_n;\,
\langle\mathbf{V}^{\ell}_j,\boldsymbol{\Delta}_{ij}\rangle_n
\right],
\label{eq:edge-invariants}
\end{equation}
where $\langle\cdot,\cdot\rangle_n$ is the normalized inner product used by
the implementation. For layer $\ell$ and head $m$,
\begin{equation}
w_{ij}^{\ell,m}
=
\operatorname{softmax}_{j\in\mathcal{N}(i)}
\operatorname{MLP}^{\ell}_{\mathrm{attn}}
\!\left(
[\boldsymbol{\phi}^{\ell}_i;
\boldsymbol{\phi}^{\ell}_j;
\mathbf{e}^{\ell}_{ij}]
\right)_m .
\label{eq:typed-attention}
\end{equation}
The vector update uses learned gates
$\mathbf{a}^{\ell,m}_{\tau}$ and $\mathbf{b}^{\ell,m}_{\tau}$ for edge
type $\tau$:
\begin{equation}
\Delta\mathbf{V}^{\ell}_i
=
\sum_m\sum_{j\in\mathcal{N}(i)}
w_{ij}^{\ell,m}
\left(
\mathbf{a}^{\ell,m}_{\tau(i,j)}
\odot\boldsymbol{\Delta}_{ij}
+
\mathbf{b}^{\ell,m}_{\tau(i,j)}
\odot\mathbf{V}^{\ell}_j
\right).
\label{eq:vector-update}
\end{equation}
Attention-weighted edge invariants are pooled across neighbors and heads
to form $\operatorname{agg}^{\ell}_i$. The scalar update is
\begin{equation}
\boldsymbol{\phi}^{\ell+1}_i
=
\boldsymbol{\phi}^{\ell}_i
+
\left(1+\boldsymbol{\gamma}_{\ell}\right)\odot
\operatorname{SwiGLU}_{\ell}
\!\left(
[\operatorname{RMSNorm}(\boldsymbol{\phi}^{\ell}_i);
\operatorname{agg}^{\ell}_i]
\right)
+
\boldsymbol{\beta}_{\ell}.
\label{eq:scalar-update}
\end{equation}
The implementation additionally RMS-normalizes vector residuals and uses
learned per-layer residual scales. Four such layers are used, with scalar
width $128$, four attention heads, three vector channels, and dropout
$0.05$.

\paragraph{Residual motion and composition.}
Only part nodes decode residual twists:
\begin{equation}
\boldsymbol{\delta}_k
=
0.2\,\tanh
\operatorname{MLP}_{\delta}
\!\left(
[\boldsymbol{\phi}_k;\operatorname{vec}(\mathbf{V}_k);
\mathbf{t}_a;\boldsymbol{\omega}_a;\boldsymbol{\iota}_a;h]
\right).
\label{eq:part-twist}
\end{equation}
Thus, the decoder receives $h$ both inside
$\boldsymbol{\iota}_a$ and as a separate scalar.
The final splat--part attention, averaged over heads, gives
$w_{ik}$ with $\sum_k w_{ik}=1$. The default Lie-algebra blend is
\begin{equation}
\boldsymbol{\delta}_i=\sum_{k=1}^{K}w_{ik}\boldsymbol{\delta}_k,
\qquad
\Delta g_i=\operatorname{Exp}(\boldsymbol{\delta}_i),
\qquad
g_i=\Delta g_i\circ g_a.
\label{eq:residual-compose}
\end{equation}
Writing $g_i=(R_i,\mathbf{t}_i)$, the Gaussian update is
\begin{equation}
\widehat{\boldsymbol{\mu}}_i
=
\mathbf{c}+R_i\mathbf{y}_i+\mathbf{t}_i,
\label{eq:mean-update}
\end{equation}
and $R_i$ is composed with the source quaternion. The last layer of the
twist decoder is initialized to zero, so the initial transformation is
exactly the supplied rigid action.

\paragraph{Appearance transport.}
On the splat $k$NN graph, local source-color transport uses
\begin{equation}
T_{ij}
=
\operatorname{softmax}_{j\in\mathcal{N}(i)}
\operatorname{MLP}_{T}
\!\left(
[\boldsymbol{\phi}_i;\boldsymbol{\phi}_j;d_{ij}]
\right),
\qquad
\widetilde{\mathbf{rgb}}_i
=
\sum_jT_{ij}\mathbf{rgb}_j.
\label{eq:appearance-transport}
\end{equation}
The appearance head predicts bounded gates. Its color residual is
\begin{equation}
\mathbf{d}_i^{\mathrm{rgb}}
=
\operatorname{clamp}_{[-0.5,0.5]}
\!\left(
\boldsymbol{\lambda}^{\mathrm{rgb}}_i\odot\mathbf{rgb}_i
+
\beta_i
\left(
\widetilde{\mathbf{rgb}}_i-\mathbf{rgb}_i
\right)
\right),
\label{eq:appearance-color}
\end{equation}
where
$\boldsymbol{\lambda}^{\mathrm{rgb}}_i\in[-0.5,0.5]^3$ and
$\beta_i\in[-0.5,0.5]$; the clamp is componentwise. The head additionally predicts
$d_i^{\alpha}\in[-3,3]$ and
$\mathbf{d}_i^{\log\mathrm{sc}}\in[-0.223,0.223]^3$. These residuals are applied as
\begin{align}
\widehat{\mathbf{rgb}}_i
&=
\operatorname{clamp}
\left(\mathbf{rgb}_i+\mathbf{d}_i^{\mathrm{rgb}},0,1\right),
&
\widehat{\alpha}_i
&=
\operatorname{sigmoid}
\left(
\operatorname{logit}
\left(\operatorname{clamp}(\alpha_i,10^{-4},1-10^{-4})\right)
+d_i^{\alpha}
\right),
\label{eq:appearance-apply-a}
\\
\widehat{\boldsymbol{sc}}_i
&=
\boldsymbol{sc}_i\odot
\exp\!\left(\mathbf{d}_i^{\log\mathrm{sc}}\right).
\label{eq:appearance-apply-b}
\end{align}
The final layers are initialized to zero, making the untrained appearance
path an identity update.

\subsection{Training Objectives}
\label{app:objectives}

\paragraph{Policy loss.}
The exact policy objective operates in normalized residual coordinates.
Let
\begin{equation}
\mathbf{u}^{(o)}_{t,h}
=
\left(
\Phi^{-1}_h\!\left(
\mathbf{a}^{(o)}_{t\rightarrow t'}\right)
-\mathbf{a}^{(o)}_{\mathrm{CV}}
\right)\oslash\boldsymbol{\sigma}
\label{eq:normalized-action-target}
\end{equation}
be the teacher residual. With dimension mask $\chi_{od}$, confidence
$q_o$, and $\epsilon=10^{-6}$, the per-actor penalty is
\begin{equation}
\ell^{(o)}_{\pi}
=
\sqrt{
\sum_{d=1}^{6}
\chi_{od}
\left(
\tanh\widehat{u}^{(o)}_{t,h,d}-u^{(o)}_{t,h,d}
\right)^2
+\epsilon^2
}.
\label{eq:policy-loss-exact}
\end{equation}
Within each sample, actor losses are averaged with normalized confidence
weights; the resulting sample losses are averaged over active samples.
This is the normalized implementation of the schematic action-space loss
in Equation~\ref{eq:policy-loss}.

\paragraph{World-model geometry losses.}
For each supervised correspondence, let $q_j\geq0$ denote its confidence.
The dense flow loss is
\begin{equation}
\mathcal{L}_{\mathrm{splat}}
=
\frac{
\sum_{o}\sum_{j\in\mathcal{J}^{(o)}_{t\rightarrow t'}}
q_j\,
\rho_W
\left(
\widehat{\mathbf{x}}_j-\mathbf{x}_j-\mathbf{f}_j
\right)}
{\sum_{o}\sum_{j\in\mathcal{J}^{(o)}_{t\rightarrow t'}}q_j},
\qquad
\rho_W(\mathbf{z})=\sum_{d=1}^{3}\sqrt{z_d^2+\epsilon^2}.
\label{eq:flow-loss-exact}
\end{equation}
The center-transport term averages predicted displacements over the same
index set $\mathcal{J}^{(o)}_{t\rightarrow t'}$ and compares that mean with
the translation $\mathbf{t}^{(o)}_a$ of the supplied target-horizon teacher
action:
\begin{equation}
\mathcal{L}_{\mathrm{center}}
=
\frac{
\sum_o q_o\,
\rho_W\!\left(
\frac{1}{|\mathcal{J}^{(o)}_{t\rightarrow t'}|}
\sum_{j\in\mathcal{J}^{(o)}_{t\rightarrow t'}}
\left(
\widehat{\mathbf{x}}_j
-\mathbf{x}_j
\right)
-\mathbf{t}^{(o)}_a
\right)}
{\sum_o q_o}.
\label{eq:center-loss-exact}
\end{equation}

\paragraph{Photometric and total loss.}
The photometric loss is the mean absolute RGB error over ground-truth
object masks:
\begin{equation}
\mathcal{L}_{\mathrm{photo}}
:=
\mathbb{E}_{o,\,p\in\mathcal{M}^{(o)}_{t'}}
\left[
\frac{1}{3}
\left\|
\widehat{\mathcal{I}}_{t'}(p)-\mathcal{I}_{t'}(p)
\right\|_1
\right].
\label{eq:photo-loss-exact}
\end{equation}
Let $\widehat{\alpha}^{(o)}(p)$ be the rendered alpha attribution of
object $o$. The implementation also adds a missing-alpha penalty
\begin{equation}
\mathcal{L}_{\mathrm{miss}}
=
\mathbb{E}_{o,\,p\in\mathcal{M}^{(o)}_{t'}}
\left[1-\widehat{\alpha}^{(o)}(p)\right].
\label{eq:missing-alpha}
\end{equation}
For the multi-horizon examples associated with one observed prefix, the world
objective is
\begin{equation}
\mathcal{L}_{W}
=
\mathbb{E}_{h}\!\left[
\mathcal{L}_{\mathrm{photo}}
+0.2\,\mathcal{L}_{\mathrm{miss}}
+0.5\,\mathcal{L}_{\mathrm{splat}}
+0.5\,\mathcal{L}_{\mathrm{center}}
\right]
+0.05\,\mathcal{L}_{\mathrm{anchor}},
\label{eq:world-loss-exact}
\end{equation}
where
$\mathcal{L}_{\mathrm{anchor}}
=\sum_{\vartheta\in\Theta_{\mathrm{geom}}}
\|\vartheta-\vartheta_0\|_2^2$
anchors geometry parameters to their initialization.

\end{document}